\documentclass[conference]{IEEEtran}
\IEEEoverridecommandlockouts
\usepackage{cite}
\usepackage{amsmath,graphics, lipsum}
\usepackage{algorithmic, algorithm}
\usepackage{array}
\usepackage{rotating}
\usepackage[table,xcdraw]{xcolor}
\usepackage{multirow}
\usepackage{lscape}
\usepackage{hyperref}
\usepackage{graphicx}
\usepackage{amssymb,cuted,mathtools,color}
\usepackage[table,xcdraw]{xcolor}
\usepackage{scalerel}
\usepackage{tikz}
\usetikzlibrary{svg.path}

\usepackage{amsmath,amssymb,amsfonts}
\usepackage{algorithmic}
\usepackage{graphicx}
\usepackage{textcomp}
\usepackage{xcolor}
\def\BibTeX{{\rm B\kern-.05em{\sc i\kern-.025em b}\kern-.08em
    T\kern-.1667em\lower.7ex\hbox{E}\kern-.125emX}}
\definecolor{orcidlogocol}{HTML}{A6CE39}
\tikzset{
  orcidlogo/.pic={
    \fill[orcidlogocol] svg{M256,128c0,70.7-57.3,128-128,128C57.3,256,0,198.7,0,128C0,57.3,57.3,0,128,0C198.7,0,256,57.3,256,128z};
    \fill[white] svg{M86.3,186.2H70.9V79.1h15.4v48.4V186.2z}
                 svg{M108.9,79.1h41.6c39.6,0,57,28.3,57,53.6c0,27.5-21.5,53.6-56.8,53.6h-41.8V79.1z M124.3,172.4h24.5c34.9,0,42.9-26.5,42.9-39.7c0-21.5-13.7-39.7-43.7-39.7h-23.7V172.4z}
                 svg{M88.7,56.8c0,5.5-4.5,10.1-10.1,10.1c-5.6,0-10.1-4.6-10.1-10.1c0-5.6,4.5-10.1,10.1-10.1C84.2,46.7,88.7,51.3,88.7,56.8z};
  }
}

\newcommand\orcidicon[1]{\href{https://orcid.org/#1}{\mbox{\scalerel*{
\begin{tikzpicture}[yscale=-1,transform shape]
\pic{orcidlogo};
\end{tikzpicture}
}{|}}}}

\begin{document}

\title{RSO: A Novel Reinforced Swarm Optimization Algorithm for Feature Selection}

\author{\IEEEauthorblockN{ Hritam Basak \orcidicon{0000-0001-5921-1230}}
\IEEEauthorblockA{\textit{Department of Electrical Engineering} \\
\textit{Jadavpur University}\\
Kolkata-700032, India\\
hritambasak48@gmail.com
}
\and
\IEEEauthorblockN{Mayukhmali Das
\orcidicon{0000-0002-9906-5167}}
\IEEEauthorblockA{\textit{Department of Electronics Engineering} \\
\textit{Jadavpur University}\\
Kolkata-700032, India \\
mayukhmalidas322@gmail.com
}
\and
\IEEEauthorblockN{Susmita Modak
\orcidicon{0000-0001-5493-7120}}
\IEEEauthorblockA{\textit{Department of Electronics Engineering} \\
\textit{Techno India University}\\
Kolkata-700091, India \\
modaksusmita2811@gmail.com
}
}
\maketitle

\begin{abstract}
Swarm optimization algorithms are widely used for feature selection before data mining and machine learning applications. The metaheuristic nature-inspired feature selection approaches are used for single-objective optimization task, though the major problem is their frequent premature convergence, leading to weak contribution to data mining. In this paper, we propose a novel feature selection algorithm named Reinforced Swarm Optimization (RSO) leveraging some of the existing problems in feature selection. This algorithm embeds the widely used Bee Swarm Optimization (BSO) algorithm along with Reinforcement Learning (RL) to maximize the reward of a superior search agent and punish the inferior ones. This hybrid optimization algorithm is more adaptive and robust with a good balance between exploitation and exploration of the search space. The proposed method is evaluated on 25 widely known UCI dataset containing a perfect blend of balanced and imbalanced data. The obtained results are compared with several other popular and recent feature selection algorithms with similar classifier configuration. The experimental outcome shows that our proposed model outperforms BSO in 22 out of 25 instances (88\%). Moreover, experimental results also show that RSO performs the best among all the methods compared in this paper in 19 out of 25 cases (76\%), establishing the superiority of our proposed method.   
\end{abstract}


\maketitle

\section{Introduction}\label{intro}

An optimization problem is the task of choosing a set of values systematically to maximize or minimize a given function with a given set of input data. More specifically, optimization is the task of selecting the "best available" of some specific objective function in a specified domain, provided that varieties of objective functions and different domains available. 
The past few years have seen an overwhelming growth of the application of single-objective and multi-objective optimization algorithms in the domain of artificial intelligence, predominantly in feature selection (FS) and is considered as the preprocessing task for several machine learning applications.
FS is the task of efficiently selecting the subset of data from larger feature sets by keeping the most relevant attributes, thereby reducing the dimensionality of the feature set, and simultaneously retaining sufficient information to perform classification of data. This is important because the irrelevant or redundant features may often lead to poor classification performance in machine learning problems with unnecessary computational cost \cite{blum1997selection}. 

In recent years, nature-inspired metaheuristic optimization algorithms, for example Particle Swarm Optimization (PSO) \cite{eberhart1995new}, Grey Wolf Optimization (GWO) \cite{mirjalili2014grey}, Genetic Algorithm (GA) \cite{schott1995fault}, Bee Swarm Optimization (BSO) \cite{karaboga2005idea} are widely used for selecting a good approximation (optimal) for various complex optimization problems, though they do not ensure the selection of the best solution always. The task of feature selection is challenging because for the original feature set of $N$ cardinality the feature selection task is to select the optimal subset among the $2^N$ intractable candidates. Hence, the number of combinations of optimal feature selection grows exponentially with the increase in the number of available features.     


In literature, different feature-selection algorithms based on nature-inspired, metaheuristic or heuristic optimization algorithms \cite{bommert2020benchmark, raj2020optimal,hu2020improved,zhang2020top} have been recently employed for different machine learning applications. Swarm-intelligence based optimization algorithms like Ant Colony Optimization \cite{ke2008efficient,engin2018new}, Particle Swarm Optimization \cite{amoozegar2018optimizing,zhang2017pso} have been modified and applied widely in recent years to serve the purpose of feature selection. Despite the efficient performance of metaheuristic feature selection algorithms over traditional machine learning approaches, the problem of an increasing amount of data makes the task difficult. Hence, researches have been made to propose hybrid optimization algorithms to improve the feature-selection performance \cite{yan2019hybrid,mafarja2017hybrid,kabir2012new,chattopadhyay2020optimizing}.     

In this paper, we propose Reinforced Swarm Optimization (RSO), a novel optimization algorithm for feature selection that incorporates the features of both reinforcement learning along with swarm intelligence based BSO algorithm. BSO \cite{karaboga2005idea}, is a metaheuristic optimization algorithm, that mimics the foraging activities of bee colony and have been used in various domains including cloud computing \cite{meshkati2019energy}, maximum satisfiability problem (MAX-SAT) \cite{djenouri2019bee}, document retrieval \cite{djenouri2018bees}, parallel computing \cite{djenouri2019exploiting}, biomedical image analysis \cite{gao2017cancer,basak2021cervical}, and many more. On the other hand, reinforcement learning (RL) has been integrated into BSO to make it more adaptive and robust powered by a suitable balance between diversification and intensification of the search space, compensating the local search of the BSO search agents.  


\section{Proposed method}\label{method}
In this section, we describe the detailed working principle of the natural bees and the inspiration behind the BSO algorithm in Section \ref{honeybee}, BSO algorithm for optimal feature selection in Section \ref{BSO}, reinforcement learning and its effects in feature selection in Section \ref{reinforcement} and the proposed RSO algorithm, incorporating the features of reinforcement learning within BSO in Section \ref{RSO}.

\subsection{Intuitive behaviour of natural bees}\label{honeybee}
Unlike other population-based methods, the BSO algorithm imitates the social hierarchy of the natural bees namely scouts, foragers, onlookers, etc. \cite{teodorovic2006bee}. 
A prospective $forager$ is a bee agent with zero information about the surrounding environment or search space and is unaware of the possible location and type of any food source or potential threat. 
The $scout$ bee, usually small in number, has the task of exploring the search space and gather information regarding the food source and pass them to the $onlooker$, who rests in the nest and processes the information collected from the foragers by implementing a probabilistic approach about the most profitable food source based on the information gathered and select the $dancers$ from numerous employed $foragers$ post advertising the information and can redefine the exploration trajectory towards the most profitable food-source. 
After collecting nectar from the food source, the forager returns to the hive and enters a decision-making process:

    1) The food source is abandoned if the remaining nectar reaches scarcity or gets completely exhausted. The employed forager bee turns into an unemployed forager; 
    2) The search can continue without additional recruiters if a sufficient amount of nectar remains in the food source;
    3) A waggle dance is performed by the forager bee to inform the nestmates about the source and the collection of nectar from the source continues.


\subsection{BSO algorithm}\label{BSO}
Bee Swarm Optimization (BSO) is a metaheuristic optimization algorithm that is inspired by the intelligent behaviour of self-organization, adaptation and hierarchical task-management of the natural bee colony. Proposed by \cite{karaboga2005idea}, the BSO algorithm is an iterative search method that solves a particular instance of optimization problem imitating the intelligent foraging behaviour and probabilistic decision-making process of natural bees to select and exploit the most profitable food source. 
Initially, the first reference solution, known as $ReferenceSol$ is generated using heuristic and is considered as the reference to determine other $N$ sets of similar solutions, together forming the $SearchSpace$. The $SearchSpace$ is defined by a set of solutions equidistant from the $ReferenceSol$ and the distance is inversely proportional to a parameter named $Flip$, which determines the convergence of the search process. Each of these solutions is considered as the starting point of local search and a bee agent is assigned to each of them. The best and fittest solution is passed to the congeners from the $Dance$ table which is further used to select the next $ReferenceSol$. The reference solutions are stored in a table named $Tab$ to avoid congestion. To avoid reaching local optima instead of a global one, the parameter $ChanceMax$ is defined carefully. It is defined as the maximum number of chance given to an artificial bee agent to explore a $ReferenceSol$ before assigning another one. If a better solution is found within the $ChanceMax$ range, intensification is done, otherwise, diversification is performed. The search stops after reaching $MaxIter$, which is the maximum number of iterations or after finding the global optima $BestSol$. The working principle BSO algorithm is explained in Algorithm \ref{algo1}.

\begin{algorithm}
    {\small
    {\em Input: Optimization problem}\\
    {\em Output: Optimal solution}\\
    Number of Search Agents: $N$\\
    Maximum number of iterations: $MaxIter$\\
    Maximum number of chaces: $ChanceMax$\\
    \begin{algorithmic}[0]
    \STATE Initialize the BSO population $X_i\:\: \forall i=1,2,3,....,N$
    \STATE $ReferenceSol$ found heuristically.
        \STATE\WHILE{t $<$ $MaxIter$}
            \STATE Insert $ReferenceSol$ into $Tab$ table
            \STATE $SearchRegion$ initialized from $ReferenceSol$
            \STATE Assigning each bee agent to every initial solution
            \STATE\FOR{each search agent}
                \STATE Local search of BSO population $X_i\:\: \forall i=1,2,3,....,N$
                \STATE Store in $Dance$ table
            \ENDFOR
            \STATE Calculate fitness of solutions found
            \STATE Best solution assigned as new $ReferenceSol$
            \STATE Update artificial bee positions.
            \STATE\IF{Optimal solution found}
                \STATE Break loop
            \ENDIF
            \STATE $t=t+1$
        \ENDWHILE
    \STATE {\em return} $BestSol$
    \end{algorithmic}
    
    \caption{Pseudo-code for Bee Swarm Optimization algorithm.}
    \label{algo1}
    }
\end{algorithm}

\subsection{Reinforcement Learning}\label{reinforcement}
Reinforced learning, also known as the Q-learning algorithm, is a machine-learning algorithm that deals with the environment with the notion of optimal cumulative reward based on the outcomes of previously implemented sets of actions. According to \cite{kaelbling1996reinforcement}, it is defined as "a way of programming agents by reward and punishment without needing to specify how the task is to be achieved". 
Let $X=\{x_1, x_2, ...., x_N \}$ be the set of states and $Y=\{y_1, y_2, ...., y_N\}$ be the set of actions, bound to select state $x_i$ from $X$. A reward $r_i$ is received for every action $a_i$ performed in set $s_i$. The algorithm tries to learn an approach to map $X\to Y$ in order to maximize the reward function which is defined by Equation \ref{eq1}.

\begin{equation}\label{eq1}
    R_{x,y}(i)=r_i+\alpha r_{i+1}+\alpha ^2 r_{i+2}+....
\end{equation}
where $\alpha$ is defined as the "discount parameter" and have range $[0,1]$. The search agents tend towards long-term rewards if the $\alpha$ value tends to 1 and short-term or immediate rewards if $\alpha$ tends towards zero. 

Temporal Difference (TD) is one of the widely used approaches in residual learning which incorporates the features of both the Monte Carlo (MC) algorithm \cite{metropolis1949monte} and Markov Decision Process (MDP) \cite{song2000optimal}. Following the original work of \cite{tesauro1995temporal}, we implement the recursive Q-learning approach, a specific TD method, to calculate the immediate reward $Q$ by acting $y_i$ in set $x_i$ given by Equation \ref{eq2}.

\begin{equation}\label{eq2}
    Q(x_i,y_i)=r(x_i,y_i)+\alpha\max Q\left(\beta(x_i,y_i), y_t \right)
\end{equation}
where $\beta(x_i,y_i)$ is the resulting state after performing action $y_i$ over set $x_i$, $y_t$ is another $t^{th}$ action. 
However, in this paper, we have slightly modified the equation to fit the purpose, given by Equation \ref{eq3}.

\begin{equation}\label{eq3}
\begin{split}
Q(x_i,y_i)=lr\times (r(x_i,y_i))+(1-lr)\times Q(x_i,y_i)\\
+\alpha\max Q\left(\beta(x_i,y_i), y_t \right)    
\end{split}
\end{equation}

where $lr$ is the learning rate and $lr\in [0,1]$. The pseudo-code of the RL algorithm is given by the Algorithm \ref{algoQ}.

\begin{algorithm}
    {\small
    \begin{algorithmic}[0]
    \STATE Initialize states $X_i\:\:\forall i=1,2,3,....N$
    \STATE Initialize actions $Y_i\:\: \forall i=1,2,3,...N$
    \STATE Initialize table elements $Q(x,y)\to 0$
        \STATE\FOR{k $\leq$ N}
            \STATE Current action $\to\:y_k$
            \STATE Current state $\to \:x_k$
            \STATE Execute $y_k$ over $x_k$
            \STATE Immediate reward $\to$ $r_k$
            \STATE New state obtained $x_t$
            \STATE $Q(x_k,y_k)\longleftarrow r_k+\alpha\max(x_t,y_t)$
            \STATE $k=k+1$
            \STATE Update $x_t\longrightarrow x_k$
        \ENDFOR
    \end{algorithmic}
    
    \caption{Pseudo-code for Reinforcement learning algorithm.}
    \label{algoQ}
    }
\end{algorithm}

\begin{table*}[htbp]
\centering
\caption{Summary of Dataset used}
\label{dataset}
\begin{tabular}{|
   c |c |c |c |c |c |c |c |}
\hline
{  \textbf{Dataset}} &
  {  \textbf{\# Attributes}} & {  \textbf{\# Instances}} & {  \textbf{\# Classes}} & {  \textbf{Dataset}} &
  {  \textbf{\# Attributes}} & {  \textbf{\# Instances}} &
  {  \textbf{\# Classes}} \\ \hline
{  Abalone} &  {  9} & {  4174} &{  28} &{  Iris} &{  4} &
  {  150} &{  2} \\ \hline
{  Australian} &{  12} &{  690} &{  2} &{  Liver} &{  7} &
  {  345} &{  2} \\ \hline
{  Biodegrade} &{  41} &{  1055} &{  2} &{  LSVT} &{  309} &{  126} &{  2} \\ \hline
{  Breastcancer} &{  9} &{  286} &{  2} &{  LungCancer} & {  56} &{  32} &{  3} \\ \hline
{  Breastcancer Wisconcin} &{  32} &{  569} &{  2} &
  {  MovementLibras} &{  90} &{  360} &{  15} \\ \hline
{  Chess} &{  6} &{  28056} &{  18} &{  Parkinson} &{  23} &{  195} &{  2} \\ \hline
{  Spect} &{  22} &{  267} &{  2} &{  Sonar} &{  60} &    {  208} &{  2} \\ \hline
{  Congress} &{  16} &{  435} &{  2} &{  Thyroid} &{  6} &
  {  215} &{  3} \\ \hline
{  Diabetes} &{  8} &{  768} &{  2} &{  Vowel} &{  10} &
  {  901} & {  15} \\ \hline
{  Glass} & {  9} & {  214} & {  7} & {  WDBC} & {  30} &
  {  569} & {  2} \\ \hline
{  Heart-C} & {  13} & {  303} & {  5} & {  Wine} & {  13} & {  178} & {  3} \\ \hline
{  Hepatitis} & {  19} & {  155} & {  2} & {  Zoo} &{  16} & {  101} & {  7} \\ \hline
{  Ionosphere} & {  34} & {  351} & {  2} & {  \textcolor{red}{}} &
  {  \textcolor{red}{}} &
  {  \textcolor{red}{}} &
  {  \textcolor{red}{}} \\ \hline
\end{tabular}
\end{table*}

\subsection{RSO: Reinforced Swarm Optimization}\label{RSO}
In this paper, we integrate Reinforcement Learning (RL) to Bee Swarm Optimization (BSO) to improve the learning process by making search agents learn from their previous experiences. One of the shortcomings of the BSO algorithm can be pointed to as the absence of intelligence or memory in their local search process which inhibits the agents to memorize the location of previously found optima. This often results in the algorithm getting stuck in local optima instead of the global one and makes the algorithm inefficient as compared to other swarm-intelligence algorithms. To address this, we propose a new algorithm by replacing the local-search algorithm with Q-learning to enable the agents' benefits from other search agents. 
In the context of FS, the inclusion or deletion of a feature set from the optimal feature subset is considered as the action whereas reward obtained is the improvement in classification accuracy and reduction in feature subset as a secondary constraint. 

In the $t^{th}$ iteration, let $Y_t=\{y_{t1},y_{t2},....,y_{tN}\}$ be the actions performed in set $X_t$ and $x_{t+1}$. The reward obtianed in set $x_t$ is obtained leveraging the classification accuracy $Acc$ and number of elements in feature subset $Num$ as follows:

\begin{equation}
\small {r_t\longleftarrow \begin{cases}
       Acc(x_t),  \text{ if } Acc(x_t)<Acc(x_{t+1})\\
        Acc(x_{t+1})-Acc(x_t),  \text{ if } Acc(x_t)>Acc(x_{t+1})\\
        \frac{Acc(x_t)}{2},  \text{ if } Num(x_t)>Num(x_{t+1})\\
       -\frac{Acc(x_t)}{2},  \text{ if } Num(x_t)<Num(x_{t+1})
        \end{cases}
        }
\end{equation}

Trivially the performance boost of the BSO method by incorporating the Residual Learning algorithm can be justified by the fact that each of the search agents learns from the previous experiences along with the experiences from other search agents. Now, in the case of the BSO algorithm, there are possibilities that one of the search agents get stuck at local minima, and considering it as the global one, the other agents converge towards that point. But in the proposed RSO method, as the agents learn from the experiences of the other search agents, the possibility of reaching the global minima increases quite significantly. 

\section{Experimental Results}\label{experiment}
The experimentations for this work was performed using Python 3.1 environment on a PC with Intel Core 7\textsuperscript{th} generation CPU and 4 GB RAM. The RSO algorithm was used for the selection of optimal feature subset followed by a classification performance using the feature subset and KNN classifier. 

\subsection{Dataset description}
To validate the performance of the proposed RSO algorithm, we have used 25 publicly available datasets from UCI machine learning repository\footnote{\url{https://archive.ics.uci.edu/ml/index.php}} and Knowledge Extraction based on Evolutionary Learning (KEEL) repository\footnote{\url{https://sci2s.ugr.es/keel/datasets.php}}. The datasets were selected while keeping a considerable diversity in the number of feature attributes, number of instances and number of classes. 
The summary of the datasets is shown in Table \ref{dataset}.

\subsection{Parameter setting }
\begin{figure*}
    \centering
    \includegraphics[width=1.55\columnwidth]{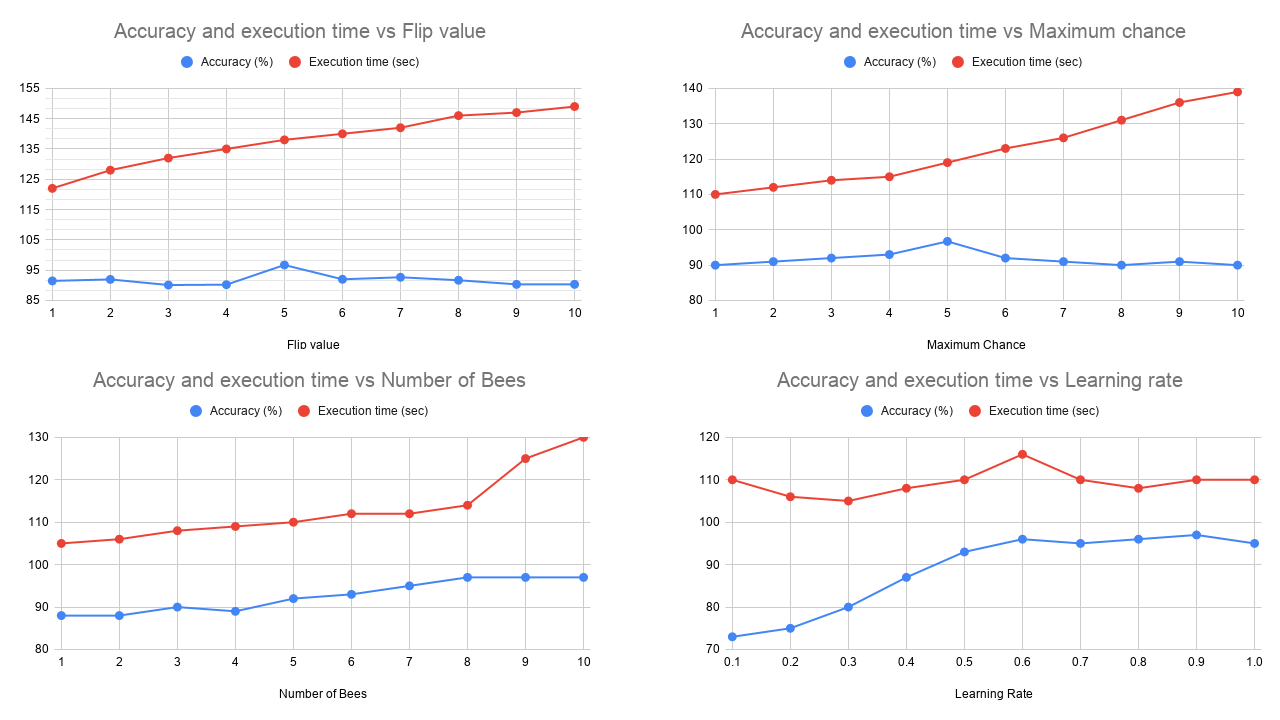}
    \caption{Analysis of classification accuracy and execution time vs. different parameters of the RSO algorithm}
    \label{expresult}
\end{figure*}

Parameter tuning has a pivotal role in the superior performance of any optimization algorithm. Hence, we have experimented with different parameters of the BSO and Residual Learning algorithm. To select the optimal set of parameters, the primary motive was to improve the classification accuracy as well as reducing the execution time. The optimal parameter setting for the algorithm was set experimentally by making a suitable compromise between these two conditions. Figure \ref{expresult} shows the experimental results with different parameter tuning of the RSO algorithm using $Ionosphere$ dataset. Different BSO parameters like $Flip$, $MaxIter$, $ChanceMax$, $NumBees$, $LsIter$ were varied with integral differences from 1 to 10 wheres different RL parameters like $lr$, $\alpha$, $\beta$ are varied within the interval from 0 to 1 with an difference of 0.1, shown in Table \ref{parameter}.

\begin{table}[h]
\centering
\caption{Optimal parameter setting for the RSO algorithm}
\label{parameter}
\begin{tabular}{|
   c | c | c |}
\hline
\textbf{Algorithm} & \textbf{Parameters} & \textbf{Optimal value} \\ \hline
& Flip  & 5  \\ \cline{2-3} 
& ChanceMax & 5 \\ \cline{2-3} 
& MaxIter  & 10  \\ \cline{2-3} 
& NumBees   & 8 \\ \cline{2-3} 
\multirow{-5}{*}{  BSO} & LsIter & 10 \\ \hline
& lr & 0.9  \\ \cline{2-3} 
& $\alpha$ & 0.2 \\ \cline{2-3} 
\multirow{-3}{*}{  RL}  & $\beta$  & 0.1           \\ \hline
\end{tabular}
\end{table}

\begin{table*}[]
\centering
\caption{Comparison of the proposed RSO method with BSO method for feature selection along with classification performance without any feature selection. The highlighted results signify the best classification accuracy for the given setting.}
\label{comp1}
\resizebox{\textwidth}{!}{%
\begin{tabular}{|
 c | c | c | c | c | c | c | c | c | c | c | c | c | c | c | c | c |}
\hline
     {     } &
  \multicolumn{4}{c|}{     {     \textbf{Without OA}}} &
  \multicolumn{6}{c|}{     {     \textbf{Proposed Method}}} &
  \multicolumn{6}{c|}{     {     \textbf{BSO}}} \\ \cline{2-17} 
\multirow{-2}{*}{     {     \textbf{Dataset}}} &
  {     Accuracy} &
  {     Precision} &
  {     Recall} &
  {     F1 Score} &
  {     Accuracy(\%)} &
  {     Precision} &
  {     Recall} &
  {     F1 Score} &
  {     \begin{tabular}[c]{@{}c@{}}Number \\ of \\ feature\end{tabular}} &
  {     Time(sec)} &
       { Accuracy(\%)} &
       { Precision} &
       { Recall} &
       { F1 Score} &
       { \begin{tabular}[c]{@{}c@{}}Number\\  of \\ feature\end{tabular}} &
       { Time(sec)} \\ \hline
{     Abalone} &
  {     19.76} &
  {     17.26} &
  {     17.27} &
  {     16.41} &
  {     \textcolor{red}{21.93}} &
  {     18} &
  {     18.88} &
  {     18.32} &
  {     7} &
  {     134} &
  {     20.33} &
  {     13.64} &
  {     13.78} &
  {     13.44} &
  {     8} &
  {     129} \\ \hline
{     Australian} &
  {     63.78} &
  {     62.5} &
  {     61.35} &
  {     61.7} &
  {     \textcolor{red}{86.96}} &
  {     86.76} &
  {     86.37} &
  {     86.54} &
  {     5} &
  {     127} &
  {     59.42} &
  {     59.21} &
  {     59.29} &
  {     59.20} &
  {     7} &
  {     132} \\ \hline
{     Biodegrade} &
  {     81.04} &
  {     79.32} &
  {     79.19} &
  {     79.8} &
  {     \textcolor{red}{84.15}} &
  {     86.93} &
  {     85.31} &
  {     86.24} &
  {     13} &
  {     146} &
  {     79.24} &
  {     77.42} &
  {     78.01} &
  {     77.68} &
  {     16} &
  {     141} \\ \hline
{     Breastcancer} &
  {     60} &
  {     58.22} &
  {     57.77} &
  {     57.8} &
  {     \textcolor{red}{97.14}} &
  {     97.67} &
  {     96.55} &
  {     97.02} &
  {     4} &
  {     121} &
  {     95.71} &
  {     96.93} &
  {     93.75} &
  {     95.08} &
  {     4} &
  {     120} \\ \hline
{     Breastcancer Wisconcin} &
  {     87.71} &
  {     87.86} &
  {     84.79} &
  {     85.99} &
  {     \textcolor{red}{95.16}} &
  {     95.29} &
  {     95.29} &
  {     95.29} &
  {     11} &
  {     139} &
  {     87.71} &
  {     87.46} &
  {     86.89} &
  {     87.14} &
  {     10} &
  {     135} \\ \hline
{     Chess} &
  {     51.27} &
  {     49.64} &
  {     46.72} &
  {     45.29} &
  {     \textcolor{red}{51.27}} &
  {     49.64} &
  {     46.72} &
  {     45.29} &
  {     6} &
  {     95} &
  {     20.20} &
  {     17.52} &
  {     16.29} &
  {     16.28} &
  {     7} &
  {     100} \\ \hline
{     Diabetes} &
  {     66.23} &
  {     61.62} &
  {     60.36} &
  {     60.65} &
  {     72.72} &
  {     69.58} &
  {     68.73} &
  {     69.11} &
  {     4} &
  {     126} &
  {     72.72} &
  {     70.91} &
  {     70.61} &
  {     70.75} &
  {     4} &
  {     124} \\ \hline
{     Glass} &
  {     74.41} &
  {     70.83} &
  {     67.59} &
  {     64.37} &
  {     \textcolor{red}{85}} &
  {     88.69} &
  {     90.28} &
  {     87.95} &
  {     7} &
  {     133} &
  {     54.54} &
  {     46.66} &
  {     43.61} &
  {     44.8} &
  {     8} &
  {     130} \\ \hline
{     Heart-C} &
  {     42.62} &
  {     31.73} &
  {     29.11} &
  {     28.98} &
  {     \textcolor{red}{58.39}} &
  {     59.66} &
  {     53.4} &
  {     55.62} &
  {     2} &
  {     130} &
  {     54.83} &
  {     24.98} &
  {     30} &
  {     26.58} &
  {     4} &
  {     129} \\ \hline
{     Hepaitits} &
  {     60} &
  {     58.92} &
  {     60} &
  {     58.33} &
  {     \textcolor{red}{78.67}} &
  {     75.96} &
  {     72.33} &
  {     74.25} &
  {     7} &
  {     135} &
  {     53.33} &
  {     50} &
  {     50} &
  {     49.76} &
  {     6} &
  {     138} \\ \hline
{     Ionosphere} &
  {     85.71} &
  {     87.07} &
  {     84.86} &
  {     85.28} &
  {     \textcolor{red}{96.72}} &
  {     91.66} &
  {     91.18} &
  {     91.31} &
  {     15} &
  {     138} &
  {     88.96} &
  {     92} &
  {     86.66} &
  {     87.95} &
  {     20} &
  {     142} \\ \hline
{     Iris} &
  {     93.33} &
  {     94.44} &
  {     93.33} &
  {     93.27} &
  {     \textcolor{red}{97.85}} &
  {     99.23} &
  {     96.19} &
  {     97.33} &
  {     2} &
  {     105} &
  {     86.64} &
  {     86.11} &
  {     86.11} &
  {     86.11} &
  {     2} &
  {     101} \\ \hline
{     LSVT} &
  {     46.15} &
  {     54.17} &
  {     54.17} &
  {     46.15} &
  {     \textcolor{red}{75.38}} &
  {     74.67} &
  {     77.81} &
  {     76.5} &
  {     213} &
  {     172} &
  {     53.84} &
  {     54.76} &
  {     55} &
  {     53.57} &
  {     224} &
  {     169} \\ \hline
{     LungCancer} &
  {     75.64} &
  {     83.33} &
  {     75} &
  {     73.33} &
  {     \textcolor{red}{98.53}} &
  {     97.5} &
  {     96.91} &
  {     97.08} &
  {     20} &
  {     157} &
  {     50} &
  {     25} &
  {     50} &
  {     33.33} &
  {     21} &
  {     158} \\ \hline
{     Parkinson} &
  {     85} &
  {     81.45} &
  {     81.88} &
  {     79.02} &
  {     93.5} &
  {     86.3} &
  {     85.79} &
  {     85.96} &
  {     12} &
  {     135} &
  {     \textcolor{red}{95}} &
  {     90} &
  {     96.8} &
  {     92.83} &
  {     16} &
  {     136} \\ \hline
{     Sonar} &
  {     78.57} &
  {     78.6} &
  {     78.41} &
  {     78.56} &
  {     \textcolor{red}{98.31}} &
  {     96.63} &
  {     96.23} &
  {     96.22} &
  {     30} &
  {     160} &
  {     85.71} &
  {     86.05} &
  {     84.72} &
  {     85.17} &
  {     42} &
  {     157} \\ \hline
{     Thyroid} &
  {     100} &
  {     100} &
  {     100} &
  {     100} &
  {     \textcolor{red}{100}} &
  {     100} &
  {     100} &
  {     100} &
  {     5} &
  {     104} &
  {     90.93} &
  {     86.92} &
  {     86.92} &
  {     86.92} &
  {     5} &
  {     99} \\ \hline
{     Wine} &
  {     60.51} &
  {     55.36} &
  {     52.89} &
  {     53.44} &
  {     97.22} &
  {     97.41} &
  {     97.22} &
  {     97.22} &
  {     9} &
  {     124} &
  {     \textcolor{red}{100}} &
  {     100} &
  {     100} &
  {     100} &
  {     10} &
  {     120} \\ \hline
{     Zoo} &
  {     95} &
  {     77.77} &
  {     83.33} &
  {     80} &
  {     \textcolor{red}{100}} &
  {     95} &
  {     96.67} &
  {     96.67} &
  {     11} &
  {     122} &
  {     100} &
  {     100} &
  {     100} &
  {     100} &
  {     11} &
  {     125} \\ \hline
{     WDBC} &
  {     77.19} &
  {     75} &
  {     75.54} &
  {     76.24} &
  {     \textcolor{red}{95.32}} &
  {     94.75} &
  {     94.75} &
  {     94.75} &
  {     14} &
  {     135} &
  {     94.73} &
  {     95.94} &
  {     93.47} &
  {     94.39} &
  {     14} &
  {     130} \\ \hline
{     Congress} &
  {     93.18} &
  {     93.37} &
  {     92.73} &
  {     93} &
  {     \textcolor{red}{97.95}} &
  {     96.54} &
  {     96.28} &
  {     96.25} &
  {     6} &
  {     127} &
  {     95.45} &
  {     93.75} &
  {     96.66} &
  {     94.94} &
  {     9} &
  {     132} \\ \hline
{     Vowel} &
  {     98.89} &
  {     98.94} &
  {     98.79} &
  {     98.89} &
  {     \textcolor{red}{100}} &
  {     100} &
  {     100} &
  {     100} &
  {     9} &
  {     126} &
  {     98.90} &
  {     99.16} &
  {     99.16} &
  {     99.13} &
  {     9} &
  {     126} \\ \hline
{     MovementLibras} &
  {     86.11} &
  {     87.56} &
  {     87.17} &
  {     86.71} &
  {     \textcolor{red}{83.33}} &
  {     86.42} &
  {     83.33} &
  {     83.52} &
  {     22} &
  {     161} &
  {     80.55} &
  {     80.35} &
  {     80.71} &
  {     77.09} &
  {     25} &
  {     164} \\ \hline
{     Liver} &
  {     62.23} &
  {     61.79} &
  {     60.33} &
  {     60.98} &
  {     \textcolor{red}{64.86}} &
  {     63.24} &
  {     65.51} &
  {     62.03} &
  {     4} &
  {     120} &
  {     62.34} &
  {     61.22} &
  {     63.59} &
  {     64.55} &
  {     5} &
  {     114} \\ \hline
{     Spect} &
  {     51.25} &
  {     50.23} &
  {     50.02} &
  {     53.66} &
  {     \textcolor{red}{77.47}} &
  {     76.33} &
  {     75.89} &
  {     76.32} &
  {     16} &
  {     129} &
  {     72.01} &
  {     70.22} &
  {     73.69} &
  {     70.66} &
  {     14} &
  {     128} \\ \hline
\end{tabular}%
}
\end{table*}
\begin{sidewaystable}

\centering
\caption{Comparison of the proposed RSO method with different existing feature selection methods for feature selection along with classification performance without any feature selection. The highlighted results signify the best classification accuracy for the given setting.}
\label{comp2}
\resizebox{\textwidth}{!}{%
\begin{tabular}{|c|c|c|c|c|c|c|c|c|c|c|c|c|c|c|c|c|c|c|c|c|c|c|c|c|c|c|c|c|}
\hline
 &
  \multicolumn{4}{c|}{Proposed method} &
  \multicolumn{4}{c|}{PSO} &
  \multicolumn{4}{c|}{MVO} &
  \multicolumn{4}{c|}{GWO} &
  \multicolumn{4}{c|}{MFO} &
  \multicolumn{4}{c|}{WOA} &
  \multicolumn{4}{c|}{HHO} \\ \cline{2-29} 
\multirow{-2}{*}{Dataset} &
  Accuracy(\%) &
  Precision &
  Recall &
  F1 Score &
  Accuracy(\%) &
  Precision &
  Recall &
  F1 Score &
  Accuracy(\%) &
  Precision &
  Recall &
  F1 Score &
  Accuracy(\%) &
  Precision &
  Recall &
  F1 Score &
  Accuracy(\%) &
  Precision &
  Recall &
  F1 Score &
  Accuracy(\%) &
  Precision &
  Recall &
  F1 Score &
  Accuracy(\%) &
  Precision &
  Recall &
  F1 Score \\ \hline
Abalone &
  {\color[HTML]{FE0000} 21.93} &
  18 &
  18.88 &
  18.32 &
  20.81 &
  15.20 &
  15.69 &
  15.05 &
  21.77 &
  12.3 &
  12.89 &
  12.48 &
  19.37 &
  10.95 &
  11.48 &
  11.07 &
  21.77 &
  12.37 &
  12.89 &
  12.48 &
  21.77 &
  12.37 &
  12.89 &
  12.48 &
  21.77 &
  12.37 &
  12.89 &
  12.48 \\ \hline
Australian &
  {\color[HTML]{FE0000} 86.96} &
  86.76 &
  86.37 &
  86.54 &
  79.71 &
  79.78 &
  80.09 &
  79.67 &
  86.95 &
  86.78 &
  86.96 &
  86.85 &
  59.42 &
  58.83 &
  58.70 &
  58.71 &
  60.86 &
  60.79 &
  60.90 &
  60.73 &
  49.27 &
  63.43 &
  53.65 &
  40.88 &
  86.95 &
  87.32 &
  86.37 &
  86.67 \\ \hline
Biodegrade &
  {\color[HTML]{FE0000} 84.15} &
  86.93 &
  85.31 &
  86.24 &
  80.18 &
  78.44 &
  79.33 &
  78.81 &
  81.13 &
  79.46 &
  80.65 &
  79.92 &
  83.01 &
  81.54 &
  81.54 &
  81.54 &
  81.13 &
  79.48 &
  79.48 &
  79.48 &
  75.47 &
  73.33 &
  72.75 &
  73.01 &
  82.07 &
  80.86 &
  83.12 &
  81.34 \\ \hline
Breastcancer &
  {\color[HTML]{FE0000} 97.14} &
  97.67 &
  96.55 &
  97.02 &
  92.85 &
  93.53 &
  90.57 &
  91.81 &
  92.85 &
  93.53 &
  90.57 &
  91.81 &
  95.71 &
  96.93 &
  93.75 &
  95.08 &
  95.71 &
  96.93 &
  93.75 &
  95.08 &
  94.28 &
  94.60 &
  92.66 &
  93.52 &
  92.85 &
  93.53 &
  90.57 &
  91.81 \\ \hline
Breastcancer Wisconcin &
  {\color[HTML]{FE0000} 95.16} &
  95.29 &
  95.29 &
  95.29 &
  89.47 &
  89.06 &
  89.06 &
  89.06 &
  89.47 &
  89.68 &
  88.36 &
  88.89 &
  87.71 &
  87.12 &
  87.59 &
  87.32 &
  89.47 &
  89.68 &
  88.36 &
  88.89 &
  94.73 &
  94.87 &
  94.18 &
  94.49 &
  94.73 &
  94.87 &
  94.18 &
  94.49 \\ \hline
Chess &
  {\color[HTML]{FE0000} 51.27} &
  49.64 &
  46.72 &
  45.29 &
  34.06 &
  34.64 &
  40.60 &
  35.54 &
  49.46 &
  46.64 &
  51.45 &
  47.93 &
  47.46 &
  30.14 &
  29.81 &
  29.91 &
  49.46 &
  46.64 &
  51.45 &
  47.93 &
  49.46 &
  46.64 &
  51.45 &
  47.93 &
  49.46 &
  46.64 &
  51.45 &
  47.93 \\ \hline
Diabetes &
  72.72 &
  69.58 &
  68.73 &
  69.11 &
  {\color[HTML]{FE0000} 76.62} &
  75.32 &
  77.15 &
  75.93 &
  {\color[HTML]{FE0000} 76.62} &
  75.71 &
  77.15 &
  75.93 &
  {\color[HTML]{FE0000} 76.62} &
  75.71 &
  77.15 &
  75.93 &
  75.32 &
  74.21 &
  75.43 &
  74.48 &
  {\color[HTML]{FE0000} 76.62} &
  75.71 &
  77.15 &
  75.93 &
  72.72 &
  70.91 &
  70.61 &
  70.75 \\ \hline
Glass &
  {\color[HTML]{FE0000} 85} &
  88.69 &
  90.28 &
  87.95 &
  59.09 &
  52.77 &
  50.27 &
  51.11 &
  59.09 &
  52 &
  45.83 &
  48.01 &
  54.54 &
  47.77 &
  43.61 &
  45.39 &
  59.09 &
  52 &
  45.83 &
  48.01 &
  54.54 &
  46.66 &
  43.61 &
  44.81 &
  54.54 &
  45 &
  43.88 &
  44.41 \\ \hline
Heart-C &
  {\color[HTML]{FE0000} 58.39} &
  59.66 &
  53.4 &
  55.62 &
  32.25 &
  16.85 &
  16.66 &
  16.25 &
  51.61 &
  24.28 &
  20.66 &
  21.09 &
  45.16 &
  15.55 &
  15.55 &
  15.55 &
  51.61 &
  28.88 &
  31.77 &
  26.82 &
  51.61 &
  21.64 &
  23.33 &
  22.14 &
  48.38 &
  35 &
  34 &
  33.56 \\ \hline
Hepaitits &
  78.67 &
  75.96 &
  72.33 &
  74.25 &
  60 &
  60.71 &
  61.11 &
  59.82 &
  {\color[HTML]{FE0000} 80} &
  80 &
  77.77 &
  78.46 &
  60 &
  58.33 &
  58.33 &
  58.33 &
  66.66 &
  69.44 &
  69.44 &
  66.66 &
  46.66 &
  47.32 &
  47.22 &
  46.42 &
  46.66 &
  44.44 &
  44.44 &
  44.44 \\ \hline
Ionosphere &
  {\color[HTML]{FE0000} 96.72} &
  91.66 &
  91.18 &
  91.31 &
  77.77 &
  86.20 &
  73.33 &
  73.81 &
  77.77 &
  86.20 &
  73.33 &
  73.81 &
  75 &
  79.46 &
  70.95 &
  71.25 &
  77.77 &
  86.20 &
  73.33 &
  73.81 &
  77.78 &
  86.20 &
  73.33 &
  73.81 &
  75 &
  85 &
  70 &
  69.74 \\ \hline
Iris &
  97.85 &
  99.23 &
  96.19 &
  97.33 &
  100 &
  100 &
  100 &
  100 &
  {\color[HTML]{FE0000} 100} &
  100 &
  100 &
  100 &
  {\color[HTML]{FE0000} 100} &
  100 &
  100 &
  100 &
  {\color[HTML]{FE0000} 100} &
  100 &
  100 &
  100 &
  {\color[HTML]{FE0000} 100} &
  100 &
  100 &
  100 &
  {\color[HTML]{FE0000} 100} &
  100 &
  100 &
  100 \\ \hline
LSVT &
  {\color[HTML]{FE0000} 75.38} &
  74.67 &
  77.81 &
  76.5 &
  69.23 &
  70.23 &
  71.25 &
  69.04 &
  69.23 &
  70.23 &
  71.25 &
  69.04 &
  69.23 &
  70.23 &
  71.25 &
  69.04 &
  61.53 &
  60.71 &
  61.25 &
  60.60 &
  69.23 &
  70.23 &
  71.25 &
  69.04 &
  61.53 &
  60.71 &
  61.25 &
  60.60 \\ \hline
LungCancer &
  {\color[HTML]{FE0000} 98.53} &
  97.5 &
  96.91 &
  97.08 &
  50 &
  50 &
  50 &
  50 &
  75 &
  83.33 &
  75 &
  73.33 &
  50 &
  50 &
  50 &
  50 &
  50 &
  25 &
  50 &
  33.33 &
  50 &
  50 &
  50 &
  50 &
  50 &
  25 &
  50 &
  33.33 \\ \hline
Parkinson &
  93.5 &
  86.3 &
  85.79 &
  85.96 &
  75 &
  67.58 &
  75 &
  68.6 &
  {\color[HTML]{FE0000} 95} &
  90 &
  96.87 &
  92.83 &
  95 &
  90 &
  96.87 &
  92.83 &
  95 &
  90 &
  96.87 &
  92.83 &
  95 &
  90 &
  96.87 &
  92.83 &
  95 &
  90 &
  96.87 &
  92.83 \\ \hline
Sonar &
  98.31 &
  96.63 &
  96.23 &
  96.22 &
  76.19 &
  78.33 &
  73.61 &
  74.07 &
  66.66 &
  66.66 &
  63.88 &
  63.70 &
  {\color[HTML]{FE0000} 100} &
  100 &
  100 &
  100 &
  90.47 &
  90.27 &
  90.27 &
  90.27 &
  76.19 &
  78.33 &
  73.61 &
  74.07 &
  71.42 &
  74.37 &
  68.05 &
  67.85 \\ \hline
Thyroid &
  {\color[HTML]{FE0000} 100} &
  100 &
  100 &
  100 &
  90.90 &
  86.92 &
  86.92 &
  86.92 &
  100 &
  100 &
  100 &
  100 &
  100 &
  100 &
  100 &
  100 &
  100 &
  100 &
  100 &
  100 &
  100 &
  100 &
  100 &
  100 &
  100 &
  100 &
  100 &
  100 \\ \hline
Wine &
  97.22 &
  97.41 &
  97.22 &
  97.22 &
  66.66 &
  62.5 &
  60.71 &
  61.32 &
  94.44 &
  95.83 &
  95.23 &
  95.21 &
  {\color[HTML]{FE0000} \textbf{100}} &
  100 &
  100 &
  100 &
  94.44 &
  93.33 &
  95.23 &
  93.73 &
  {\color[HTML]{FE0000} 100} &
  100 &
  100 &
  100 &
  {\color[HTML]{FE0000} 100} &
  100 &
  100 &
  100 \\ \hline
Zoo &
  {\color[HTML]{FE0000} 100} &
  95 &
  96.67 &
  96.67 &
  100 &
  100 &
  100 &
  100 &
  90.90 &
  66.66 &
  66.66 &
  66.66 &
  90.90 &
  66.66 &
  66.66 &
  66.66 &
  100 &
  100 &
  100 &
  100 &
  90.90 &
  66.66 &
  66.66 &
  66.66 &
  90.90 &
  66.66 &
  66.66 &
  66.66 \\ \hline
WDBC &
  {\color[HTML]{FE0000} 95.32} &
  94.75 &
  94.75 &
  94.75 &
  87.71 &
  89.52 &
  85.48 &
  86.66 &
  87.71 &
  87.46 &
  86.89 &
  87.14 &
  78.94 &
  78.94 &
  76.72 &
  77.38 &
  89.47 &
  89.68 &
  88.36 &
  88.89 &
  84.21 &
  83.76 &
  83.24 &
  83.47 &
  91.22 &
  92.09 &
  89.83 &
  90.66 \\ \hline
Congress &
  {\color[HTML]{FE0000} 97.95} &
  96.54 &
  96.28 &
  96.25 &
  90.90 &
  88.83 &
  91.42 &
  89.88 &
  90.90 &
  88.88 &
  93.33 &
  90.17 &
  95.45 &
  93.75 &
  96.66 &
  94.94 &
  97.72 &
  96.66 &
  98.33 &
  97.42 &
  97.72 &
  96.66 &
  98.33 &
  97.42 &
  90.90 &
  88.88 &
  93.33 &
  90.17 \\ \hline
Vowel &
  {\color[HTML]{FE0000} 100} &
  100 &
  100 &
  100 &
  97.80 &
  95.83 &
  98.05 &
  96.54 &
  97.80 &
  95.83 &
  98.05 &
  96.54 &
  96.70 &
  97.46 &
  97.08 &
  97.06 &
  97.80 &
  98.16 &
  97.91 &
  97.93 &
  97.80 &
  98.16 &
  97.91 &
  97.93 &
  97.80 &
  95.83 &
  98.05 &
  96.54 \\ \hline
MovementLibras &
  {\color[HTML]{FE0000} 83.33} &
  86.42 &
  83.33 &
  83.52 &
  80.55 &
  80.35 &
  80.71 &
  77.09 &
  83.33 &
  81.30 &
  84.28 &
  79.99 &
  83.33 &
  82.14 &
  82.5 &
  79.47 &
  83.33 &
  81.66 &
  84.28 &
  79.86 &
  80.55 &
  81.76 &
  80.71 &
  77.73 &
  77.77 &
  80.65 &
  78.33 &
  76.16 \\ \hline
Liver &
  {\color[HTML]{FE0000} 64.86} &
  63.24 &
  65.51 &
  62.03 &
  61.25 &
  60.94 &
  61.27 &
  61.11 &
  62.66 &
  62.47 &
  63.08 &
  62.69 &
  63.52 &
  62.63 &
  63.02 &
  62.94 &
  62.66 &
  62.47 &
  63.08 &
  62.69 &
  60.78 &
  61.29 &
  60.36 &
  63.52 &
  62.63 &
  63.02 &
  62.94 &
  62.66 \\ \hline
Spect &
  {\color[HTML]{FE0000} 77.47} &
  76.33 &
  75.89 &
  76.32 &
  75.88 &
  76.36 &
  76.55 &
  76.48 &
  75.88 &
  76.36 &
  76.55 &
  76.48 &
  75.88 &
  76.36 &
  76.55 &
  76.48 &
  76.15 &
  75.89 &
  76.32 &
  76.06 &
  75 &
  75 &
  75 &
  75 &
  75.88 &
  76.36 &
  76.55 &
  76.48 \\ \hline
\end{tabular}%
}
\end{sidewaystable}

\subsection{Performance evaluation }
The performance of the proposed method was evaluated on 25 standard datasets where we have selected optimal feature subset using RSO followed by classification using KNN classifier. The classification performance was evaluated using the evaluation metrics given by Equation \ref{acc}-\ref{f1}.
\begin{equation}\label{acc}
    Accuracy=\frac{TP+TN}{TP+TN+FP+FN}
\end{equation}
\begin{equation}
    Precision=\frac{TP}{TP+FP}
\end{equation}
\begin{equation}
    Recall=\frac{TP}{TP+FN}
\end{equation}
\begin{equation}\label{f1}
    F1 score=\frac{2}{\frac{1}{Precision}+\frac{1}{Recall}}
\end{equation}

where $TP$ = True Positive, $FP$ = False Positive, $TN$ = True Negative, and $FN$ = False Negative.

\subsection{Comparison with existing methods}
We have evaluated the performance of the proposed RSO method with different existing optimization algorithms. Table \ref{comp1} shows the comparison of the experimental results in terms of \textbf{accuracy, precision, recall, F1 score, execution time, and the number of selected features}, obtained from our proposed method with the same obtained from the BSO algorithm.
It is evident from the table that our proposed RSO method outperforms the BSO algorithm in 22 out of the 25 cases in terms of classification accuracy, by using a significantly smaller subset of feature data and thereby reducing the execution time.


We have also compared the obtained results with several feature selection algorithms like Particle Swarm Optimization (PSO) \cite{eberhart1995new}, Grey Wolf Optimization (GWO) \cite{mirjalili2014grey}, Genetic Algorithm (GA) \cite{schott1995fault}, Harris Hawk Optimization (HHO) \cite{heidari2019harris}, Multi-Verse Optimization (MVO) \cite{mirjalili2016multi}, Moth Flame Optimization \cite{mirjalili2015moth}, Whale Optimization Algorithm (WOA) \cite{mirjalili2016whale} as shown in Table \ref{comp2}. The proposed method outperformed all the methods compared in this paper in 19 out of 25 cases in terms of fitness of selected features, which is reflected in classification accuracy. However, our model performed inferior in the case of $Wine$ dataset with the best classification accuracy of 97.22\% whereas most of the other methods were able to produce a superior feature subset, resulting in a classification accuracy of 100\%. In the case of $Diabetes$ dataset, PSO, MVO, WOA and GWO performed the best with a classification accuracy of 76.62\% as compared to 72.72\% from the proposed RSO method. The MVO method performed the best in $Hepatitis$ dataset as compared to 78.67\% from our proposed method. For $Iris$ dataset, all the methods produced a classification accuracy of 100\% whereas the RSO method produced a result of 97.85\%. MVO, MFO, WOA, and HHO produced similar results to the BSO method in $Parkinson$ dataset with a classification accuracy of 95\% as compared to 93.5\% of our proposed method. In the case of $Sonar$ data, the GWO performs the best with a classification accuracy of 100\% as compared to 98.31\% from RSO. Our proposed method performs the best in all other datasets as shown in Table \ref{comp2}.

\section{Conclusion and future work}\label{conclusion}
In this paper, we propose a new hybrid wrapper-based feature selection algorithm named RSO, which integrates the Residual Learning algorithm with the metaheuristic BSO algorithm. Experimental results show that our proposed method outperforms the BSO as well as the existing and popularly known metaheuristic optimization algorithms in feature selection task in terms of accuracy by selecting comparatively fewer optimal features. In future, we plan to extend our research by experimenting and observing the performance of different hybrid optimization algorithms. We also plan to observe the performance of RSO on deep features, to study the impact of RL in the performance of feature selection algorithms.

\bibliographystyle{IEEEtran}
\bibliography{References}

\end{document}